\documentclass{article} %[10pt,twocolumn,letterpaper]

\date{}

\usepackage[a4paper, total={6in, 10in}]{geometry}
\usepackage{times}
\usepackage{epsfig}
\usepackage{graphicx}
\usepackage{amsmath}
\usepackage{amssymb}

\usepackage{amsfonts}
\usepackage{multirow}
\usepackage{subfigure}
\usepackage{array}
\usepackage{mathrsfs}
\usepackage{bm}
\usepackage{amsthm}
\usepackage{multirow}
\usepackage{lineno}
\usepackage{url}

\usepackage{algorithmic}
\usepackage{algorithm}

\usepackage[capitalize,noabbrev]{cleveref}
\usepackage{booktabs}
\usepackage[numbers,sort&compress]{natbib}

\begin{document}
%\linenumbers

%%%%%%%%% TITLE
\title{A Molecular Multimodal Foundation Model Associating Molecule Graphs with Natural Language}
% A Multimodal Chemical Foundation Model Enabling AI to Imagine Molecule Graphs from Natural Language
% A Chemical Multimodal Foundation Model Bridging Molecule Graphs and Natural Language

\author{Bing Su$^{1,2}$, Dazhao Du$^{3,4}$, Zhao Yang$^{1,2}$, Yujie Zhou$^{1,2}$, Jiangmeng Li$^{3,4}$,\\ Anyi Rao$^{5}$, Hao Sun$^{1,2}$, Zhiwu Lu$^{1,2}$, Ji-Rong Wen$^{1,2}$\\
\\
$^1$ Gaoling School of Artificial Intelligence,\\
Renmin University of China, Beijing 100872, China\\
$^2$ Beijing Key Laboratory of Big Data Management and Analysis Methods\\
$^3$ Science \& Technology on Integrated Information System Laboratory,\\
Institute of Software, Chinese Academy of Sciences, Beijing, China\\
$^4$ University of Chinese Academy of Sciences, Beijing, China\\
$^5$ The Chinese University of Hong Kong, Hong Kong, China\\
% $^*$Corresponding author: Ji-Rong Wen, {\tt\small E-mail: jrwen@ruc.edu.cn}
% For a paper whose authors are all at the same institution,subingats@gmail.com;
%\and
}

\maketitle
%\ificcvfinal\thispagestyle{empty}\fi

%%%%%%%%% ABSTRACT
\begin{abstract}
Although artificial intelligence (AI) has made significant progress in understanding molecules in a wide range of fields, existing models generally acquire the single cognitive ability from the single molecular modality. Since the hierarchy of molecular knowledge is profound, even humans learn from different modalities including both intuitive diagrams and professional texts to assist their understanding. Inspired by this, we propose a molecular multimodal foundation model which is pretrained from molecular graphs and their semantically related textual data (crawled from published Scientific Citation Index papers) via contrastive learning. This AI model represents a critical attempt that directly bridges molecular graphs and natural language. Importantly, through capturing the specific and complementary information of the two modalities, our proposed model can better grasp molecular expertise. Experimental results show that our model not only exhibits promising performance in cross-modal tasks such as cross-modal retrieval and molecule caption, but also enhances molecular property prediction and possesses capability to generate meaningful molecular graphs from natural language descriptions. We believe that our model would have a broad impact on AI-empowered fields across disciplines such as biology, chemistry, materials, environment, and medicine, among others. %In particular, our model possesses powerful inspiration and imagination to generate meaningful molecular graphs from natural language descriptions, which has great potential in materials synthesis and drug discovery.powerful inspiration and
\end{abstract}

%%%%%%%%% BODY TEXT
%\section{Introduction}
\vspace{0.2in}

Understanding molecule-related knowledge and discovering molecular properties are critical for scientific exploration across a wide range of disciplines such as biomedicine, chemistry, materials, etc. Traditional methods require professional skills to conduct large amounts of trial-and-error wet biochemical experiments~\cite{hajduk2007decade,clark2009design,rodrigues2016counting}, which are not only expensive but also time-consuming. With the advancement of deep learning~\cite{lecun2015deep}, using AI to assist in scientific exploration such as predicting molecular properties and generating molecule candidates has become possible witnessed by many critical signs of progress~\cite{lu2018accelerated,jin2020multi,walters2020applications,kotsias2020direct,mahmood2021masked}.%and functions is for various applications in these areas., discovering new molecules with desired characteristics the foundation for scientific exploration hypothesize or design molecules based on experience or inspiration, and with preferred properties for gaining such knowledge and properties

\begin{figure}[t]
\centering
\includegraphics[width=0.95\columnwidth]{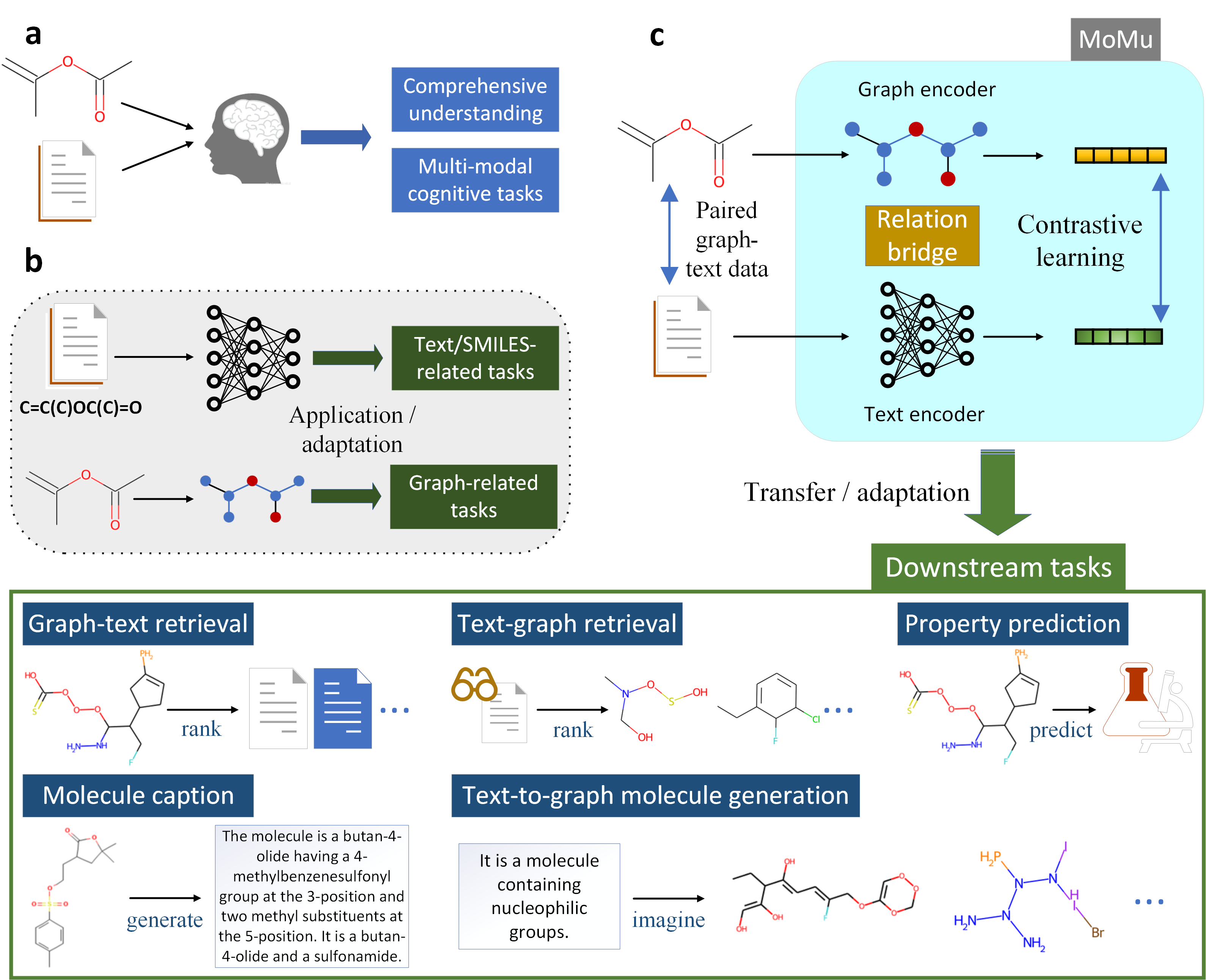}
\caption{Conceptual comparison of our MoMu model with the human brain and existing single-modal AI models. (a) The human brain can comprehensively understand molecular knowledge by learning from multiple modalities. (b) Existing AI models generally employ a single network to gain a single cognitive ability from a single modality of molecules. These models mainly fall into two categories: (top) language-based models take as input natural language texts and/or SMILES strings, which can only be applied to text-related tasks; (bottom) graph-based models take molecular graphs as input, which can only be adapted to graph-related tasks. (c) Our MoMu model learns from weakly-correlated paired text-graph data to associate the molecular graph modality with the natural language modality. It consists of two encoders to tackle the two modalities, respectively, which are jointly trained via contrastive learning. Due to the strong generalization ability of the learned representations, MoMu can be adapted to various downstream tasks such as cross-modality retrieval, molecule caption, property prediction, and text-to-graph molecule generation, and thus effectively facilitates molecular-related scientific exploration.}
\label{fig:overall}
%\vskip -0.2in
\end{figure}

Different from humans who understand molecules from multiple modalities as shown in Figure~\ref{fig:overall}(a), most existing AI models are specified for a single cognitive ability (e.g., property prediction, molecule generation, or literature comprehension) and a single modality of molecules (e.g., molecular graph, SMILES string, or text), as shown in Figure~\ref{fig:overall}(b). These models are mainly divided into two categories. Firstly, language-based models take natural language about molecular knowledge and/or SMILES strings as input. For example, in~\cite{wang2019smiles,chithrananda2020chemberta}, molecular property prediction models are designed for SMILES strings of molecules; several works~\cite{beltagy2019scibert,li2019biomedical,lee2020biobert} focus on literature comprehension learning from biochemical texts; in generation models~\cite{kusner2017grammar,dai2018syntax,segler2018generating,gomez2018automatic,honda2019smiles,popova2019molecularrnn,wang2021multi,guo2022improving,flam2022language,hoffman2022optimizing}, the generated molecules are represented as SMILES. Secondly, graph-based models can only process molecular graphs. For example, in~\cite{hy2018predicting,coley2019graph,zaidi2022pre,li2022glam,wang2021molclr,wang2021chemical,chen2021algebraic}, graph neural network (GNN)~\cite{gilmer2017neural} based molecular property prediction models are learned from molecular graphs; in~\cite{jin2018junction,shi2019graphaf,zang2020moflow,ma2021gf,luo2021graphdf,chen2021deep,lee2022mgcvae}, generation models are learned from graph data to directly generate molecular graphs. Note that extensive manual annotations or collections of specific properties are required for training these models, while other properties of molecules and associated conditions are often omitted and thus cannot be perceived by the models. Therefore, they can only deal with one modality of the molecule and cannot acquire comprehensive molecule understanding.

Professional molecular expertise is usually documented in published papers or books in natural language. Human beings can learn such knowledge by reading the related literature. There have been some works~\cite{beltagy2019scibert,lee2020biobert} toward making AI models acquire such knowledge from a large number of professional literature. However, learning knowledge only at the language level may simply associate entities of molecule names with property descriptions, resulting in the challenge in connecting property descriptions with molecular structures. Recently, a deep learning system is developed~\cite{zeng2022deep} to jointly learn molecule-related texts and molecular SMILES strings to establish their relationships (as also included in the top of Figure~\ref{fig:overall}(b)). However, this approach has distinct drawbacks: one-dimensional SMILES strings may lose some molecular structure information and cannot capture structural similarities between molecules; moreover, the SMILES of a molecule is not unique.

Different from molecular SMILES strings, molecule graphs are more intuitive and can better reveal the functional structures. Since the molecular knowledge is profound and difficult to understand, even human beings assist their learning process by associating natural language descriptions with molecule graphs. Inspired by this, we propose a molecular multi-modal foundation model (MoMu) that consists of two separate encoders respectively for molecule graphs and texts, as shown in Figure~\ref{fig:overall}(c). To train the model, we collect about 15K paired molecule graph-text data, where the text of a molecule is retrieved from the SCI paper dataset~\cite{s2ocr}. The two encoders are jointly trained by contrastive learning so that representations of molecule graphs are as similar as representations of their related texts and as dissimilar as those of unrelated texts. In this way, our MoMu model is able to associate molecule graphs with their biomedical text descriptions. MoMu represents a critical attempt to bridge the two modalities. %To our best knowledge, this is the first AI model that bridges the two modalities. % molecule graphs and natural languages. %bridge molecular graphs and natural language.% by using the molecule name as the query We propose a molecular multi-modal foundation model that

Recently, several multi-modal foundation models~\cite{radford2021learning,jia2021scaling,li2020oscar,fei2022towards} have been developed to connect images and texts. These models are trained with large-scale weakly-related image-text data. Since it is more difficult for humans to learn specialized biological and chemistry literature than general language knowledge, making AI acquire multi-modal molecular knowledge also faces greater challenges. Firstly, much fewer paired molecular graph-text data are available than more general image-text data. Secondly, learning professional molecular knowledge requires basic cognitive abilities, and thus it is difficult to learn a molecular multi-modal model from scratch. To tackle these challenges, we employ the molecular graph model~\cite{you2020graph} and biomedical text model~\cite{beltagy2019scibert,zeng2022deep} pre-trained from large-scale single-modality unlabeled data as the initialization of the two encoders in our foundation model, and finetune them with our collected professional graph-text data through contrastive learning. As a result, our foundation model requires much less manual labor and computational resources for training.

By processing molecules in multiple modalities, our foundation model can be applied to a wide range of downstream tasks. For cross-modality tasks, experimental results show that our model achieves a significant improvement in cross-modality retrieval and improves molecule captioning by incorporating the graph feature. More importantly, our model implicitly establishes the connection between structures of molecules and language descriptions, thereby allowing text-to-graph generation. In many realistic applications such as catalyst design and targeted drug discovery, designing new molecules often relies on the knowledge, experience, and inspiration of experts. Although deep learning-based generation models~\cite{jin2018junction,shi2019graphaf,zang2020moflow,ma2021gf,lee2022mgcvae} have been developed, they rely on large numbers of molecules with specified properties for modeling training. Differently, based on our MoMu model, we design a zero-shot molecule generation method that can directly imagine new molecules from textual descriptions of the desired conditions. For the graph modality, we apply the pre-trained MoMu to molecular property prediction and find that our graph encoder supervised by the weakly related language descriptions outperforms other self-supervised methods trained only from molecule graphs. These findings demonstrate that, due to the strong generalization and imagination abilities, our pre-trained MoMu model can advance scientific exploration and thus has the potential to lead to broad impact in biology, chemistry, materials, medicine, and other molecular-related fields.

%tasks such as cross-modal In cross-modality molecule/text retrieval and molecule caption,graph encoder of Different labeled molecule data are required to retrain different models for different properties and demands. This is not only labour and resource-consuming, but also difficult to tackle situations when multiple conditions need to be satisfied simultaneously.the language-based method with specified properties

%The major contributions of this paper can be summarized as follows.
%1. We propose a molecular multi-modal foundation model that associates molecule graphs with their biomedical text descriptions. To our knowledge, our model is the first to bridge molecule graphs and natural languages.
%2. We show the applications of our model in a wide range of downstream tasks, including graph-related tasks such as property prediction and cross-modality tasks such as cross-modal retrieval and molecule caption.
%3. We show that our model can be employed to design new molecules according to text descriptions in a zero-shot manner. %, which may have a significant impact on material and drug design.

\section*{Results}

Our molecular multimodal foundation model, MoMu, is pre-trained based on the paired multi-modal data consisting of molecule graphs and their weakly-related biochemical descriptions retrieved from publicly available Scientific Citation Index (SCI) papers. The pre-trained MoMu exhibits strong generalization abilities in a wide range of downstream tasks, including cross-modal molecule retrieval, molecule caption, zero-shot molecule generation, and molecular property prediction, as shown in this section.

\subsection*{Data collection and overview of MoMu}
We construct a dataset of molecular graph-text pairs for pre-training our model. We first collect the names, synonyms, and SMILES strings of the top 50K molecular compounds in PubChem~\cite{pubchem}. To obtain the molecule graphs of the collected compounds, we employ the smiles2graph function provided by OGB~\cite{ogb}, which converts SMILES strings into molecular graphs. As shown in the Supplementary Figure 1(a), for each molecule, we retrieve related texts in published scientific papers in the S2orc~\cite{s2ocr} database into a document as weak semantic supervision by using its name as the query. S2orc is a corpus database containing 136M+ papers from different fields, and we only retrieve papers in the fields of Medicine, Biology, Chemistry, and Computer Science, as they are more likely to contain molecule-relevant descriptions. Finally, we obtain 15,613 graph-document pairs to form our multi-modal molecular dataset. There are about 37 million paragraphs in all the collected documents. %Details of the data collection are presented in the Method section.

%{\bf Overview of the MoMu model.}
The overall architecture and the pre-training process of our MoMu model are illustrated in Figure~\ref{fig:overall}(c) and the Supplementary Figure 1(b). MoMu consists of a text encoder and a graph encoder, which encode the molecular graphs and texts into a joint representation space, respectively. We use Graph Isomorphism Network (GIN) \cite{xu2018powerful} and Bert~\cite{devlin2018bert} as the graph and text encoders, respectively. We train MoMu based on our collected paired dataset. For each pair within a minibatch, we utilize two different types of graph augmentations to create two separate graphs from the molecular graph, and randomly sample two different sentences from the document. Following the contrastive paradigm in DeClip~\cite{li2021supervision}, we use inter-modal and intra-modal contrastive learning~\cite{chen2020simple,he2020momentum,radford2021learning} as the proxy task for pretraining, which employs the InfoNCE loss~\cite{hjelm2018learning}. The proxy task aims to bring samples from different modalities with the same semantic information closer together in the feature space while pushing samples with different semantics away. %To enhance the learning of the graph model, we further employ contrastive learning within the graph modalities, i.e., pulling the features of two enhanced graphs from the same molecule closer in the feature space.% and employ the widely-used language model as the text encoder As a result, each modality has two samples that contain the same semantic information.

Compared with general image-text data, there are relatively much fewer molecule-related graph-text data, which is not sufficient to train graph and text encoders from scratch. Just as human beings should also have the ability to recognize graphs and languages when learning specialized knowledge, enabling AI to learn specialized molecular knowledge also needs to be founded on trained general graph and text encoders. Therefore, we initialize the graph encoder with the self-supervised trained weights of GIN provided in~\cite{you2020graph}, and initialize the text encoder with the pre-trained weights of BERT provided by Sci-BERT~\cite{beltagy2019scibert} and KV-PLM~\cite{zeng2022deep}, respectively. We denote MoMu initialized with the weights of Sci-BERT and KV-PLM by MoMu-S and MoMu-K, respectively. Details of the data collection and pretraining method for MoMu are presented in the Methods section. The pre-trained MoMu is able to process molecular graphs and natural language texts in a unified manner, and acquire the common and transferable knowledge from such heterogeneous data that can be easily generalized to different downstream tasks.

\begin{figure}[t!]
\centering
\includegraphics[width=0.95\columnwidth]{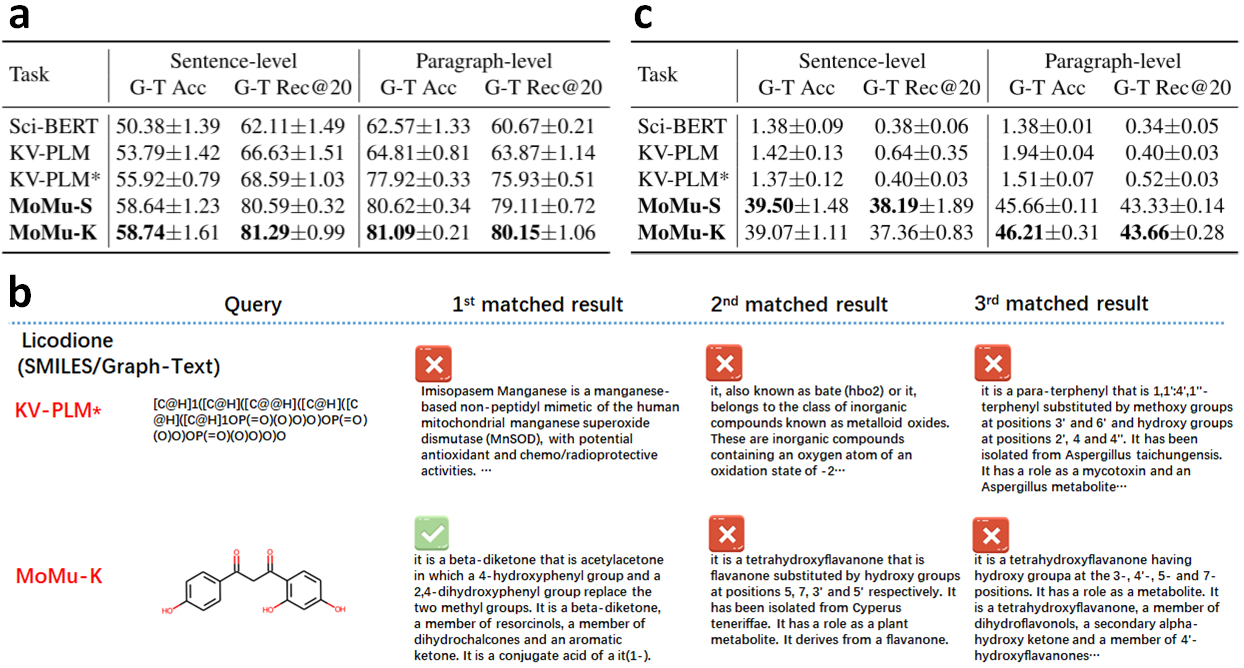}
\caption{Graph-to-text retrieval results. (a) The performance of graph-to-text (G-T) retrieval on the PCdes dataset, where the results of the compared methods for the sentence-level retrieval are reported in \cite{zeng2022deep}. (b) Retrieval results by using an example/SMILES in the test set of PCdes as the query by KV-PLM* and our MoMu-K. (c) The performance of zero-shot graph-to-text (G-T) retrieval on our collected test set.}
\label{fig:G2TRetrieval}
%\vskip -0.2in
\end{figure}

\subsection*{Cross-modality retrieval}
Since our MoMu model is pre-trained by matching weakly-correlated texts to corresponding molecular graphs, it is able to process both the graph and text modalities of molecules. We evaluate its performance in cross-modality retrieval. Given a molecule graph, graph-to-text (G-T) retrieval aims to retrieve the most relevant text descriptions of this molecule. Conversely, given a text paragraph, text-to-graph (T-G) retrieval aims at retrieving the most relevant molecule graph it describes. We evaluate MoMu on the PCdes dataset~\cite{zeng2022deep}, which contains SMILES and the paired property descriptions of 15K molecules in PubChem. The dataset has been divided into a training set of 10,500 pairs, a validation set of 1,500 pairs, and a test set of 3,000 pairs\footnote{Two SMILES in the test set cannot be converted into graphs by Rdkit so we use the remaining 2998 pairs for testing.}. We convert the SMILES string in each pair into the molecule graph. In the G-T/T-G task, we calculate and rank the cosine similarities between the representation of the query graph/text by the graph/text encoder of MoMu and the representations of all texts/graphs by the text/graph encoder of MoMu, as shown in the Supplementary Figure 2. Following~\cite{zeng2022deep}, we perform retrieval in randomly sampled mini-batches (64 pairs per batch) and all test pairs, respectively, and report the average accuracy of the top-1 retrieval result and the recall of the top-20 results, respectively. In~\cite{zeng2022deep}, one sentence is randomly sampled from the text corresponding to each molecule for retrieval, and we denote this setting by sentence-level retrieval. We further evaluate the setting of using the full description paragraph per molecule, which is denoted by paragraph-level retrieval.

As in~\cite{zeng2022deep}, the compared methods including Sci-BERT~\cite{beltagy2019scibert}, and KV-PLM~\cite{zeng2022deep}\footnote{KV-PLM* differs from KV-PLM by processing SMILES with double tokenizers. The text encoder of MoMu-K is initialized from KV-PLM*.}, as well as our MoMu, are fine-tuned on the training set of PCdes to make a fair comparison. For the G-T task, comparisons of different methods are shown in Figure~\ref{fig:G2TRetrieval}(a). Both MoMu-S and MoMu-K outperform other methods that directly use SMILES for retrieval. We illustrate a retrieval example in Figure~\ref{fig:G2TRetrieval}(b) by using the licodione molecule as the query. None of the top-3 paragraph-level retrieval results of KV-PLM* from the whole test set of PCdes match the SMILES of this molecule, while the top-1 result of our MoMu-K accurately hits the corresponding description of the molecular graph. To verify the generalization of MoMu, we conduct zero-shot retrieval, where encoders of the pre-trained MoMu are directly applied without fine-tuning. Considering that some of these 15K pairs in PCdes may have been collected as our pre-training data for MoMu, we collect 5,562 graph-text pairs with compound ids ranging from 50,000 to 100,000 from PubChem that are not used in pre-training. Comparisons with Sci-BERT and KV-PLM on this collected test set for zero-shot retrieval are shown in Figure~\ref{fig:G2TRetrieval}(c). MoMu also outperforms Sci-BERT and KV-PLM significantly, which further demonstrates the generalization ability of MoMu.

\begin{figure}[t!]
\centering
\includegraphics[width=0.95\columnwidth]{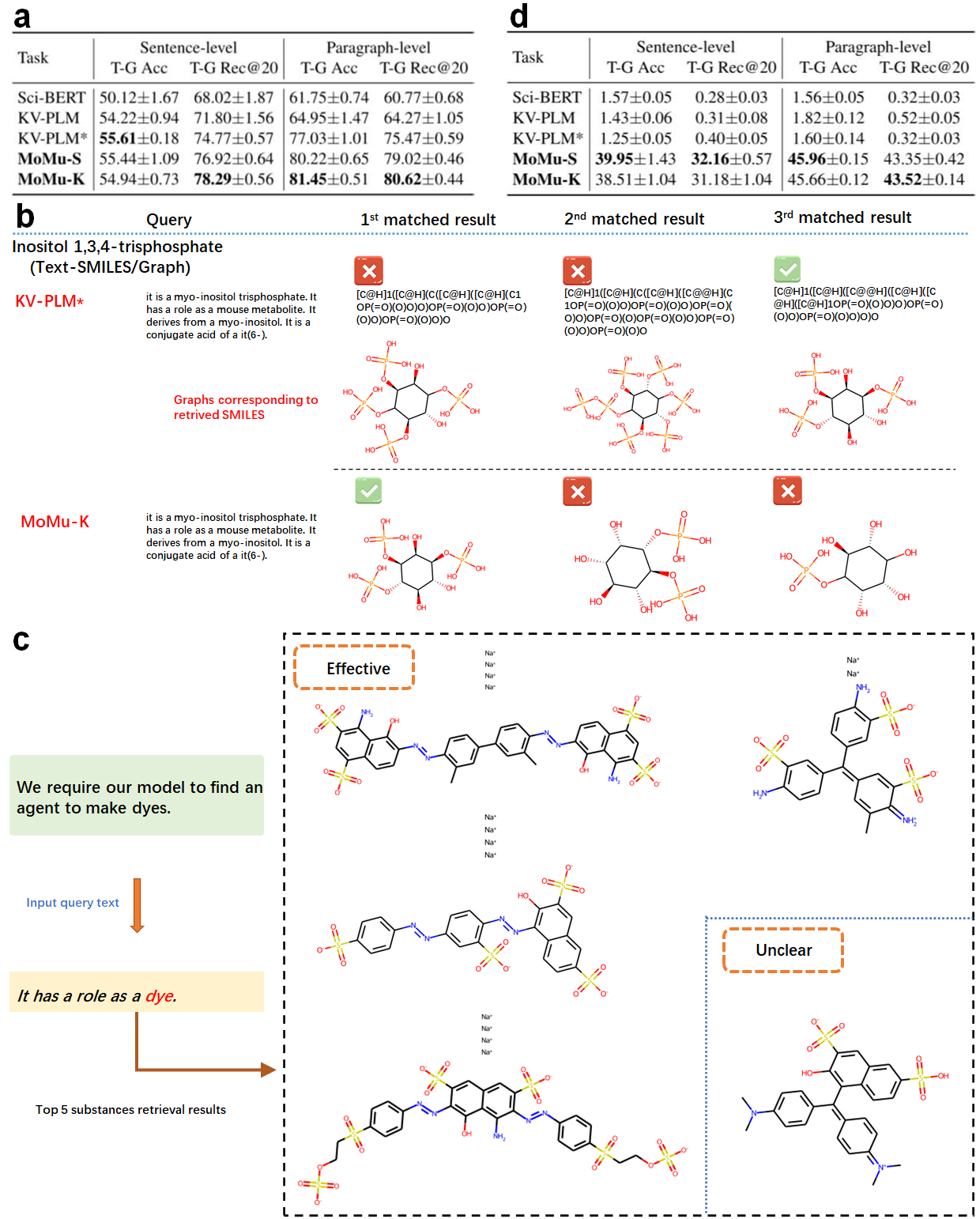}
\caption{Text-to-graph retrieval results. (a) The performance of text-to-graph (T-G) retrieval on the PCdes dataset, where the results of the compared methods for the sentence-level retrieval are reported in \cite{zeng2022deep}. (b) Retrieval results by using a text paragraph in the test set of PCdes as the query by KV-PLM* and our MoMu-K. (c) A case study by using a query text to retrieve molecules that can make dyes. Four of the top-5 molecules retrieved by MoMu-K are confirmed to be effective. (d) The performance of zero-shot text-to-graph (T-G) retrieval on our collected test set.}
\label{fig:T2GRetrieval}
%\vskip -0.2in
\end{figure}

For the T-G task, comparisons of different methods are shown in Figure~\ref{fig:T2GRetrieval}(a). Both MoMu-S and MoMu-K generally outperform other methods. We also illustrate a retrieval example in Figure~\ref{fig:T2GRetrieval}(b) by using a paragraph as the query to retrieve SMILES or graphs from the whole test set of PCdes. The third molecule in the ranking list of KV-PLM* matches the groundtruth, while the top-1 result of MoMu-K hits and the other retrieved molecules are also similar to the groundtruth. In Figure~\ref{fig:T2GRetrieval}(c), we further conduct a case study to discovery dye molecules from the test set of PCdes by our MoMu-K. We use the text ``It has a role as a dye.'' as the query. In the top-5 retrieved molecules, four are confirmed to be valid dyes. It is unclear whether the remaining molecule can be used as a dye, because there is no record of dye properties in PubChem for this molecule. Comparisons with Sci-BERT and KV-PLM on our collected test set for zero-shot retrieval are shown in Figure~\ref{fig:T2GRetrieval}(d). Again, our method outperforms other methods significantly. These results demonstrate that, compared with KV-PLM which jointly models SMILES of molecules and language texts, our MoMu can better bridge molecular structures and natural language descriptions.

On both datasets with different settings, MoMu-S and MoMu-K obtain comparable results, i.e., initializing the text encoder of MoMu with KV-PLM does not lead to better performance than Sci-BERT. This shows that the structural information learned from one-dimensional SMILES strings of molecules can not be easily transferred to structured molecular graphs, while MoMu directly employs the graph neural network to capture structural information with the supervision of language descriptions.%Especially for the G-T task, MoMus outperform KV-PLM by a significant margin of about 10\%.RXNFP~\cite{rxnfp}, BERT$_{wo}$~\cite{devlin2018bert}, SMI-BERT~\cite{zeng2022deep},generally
%As shown in Table~\ref{tab:retrieval}, zero-shot MoMus archive better results than baselines that are not specifically adapted for molecular modalities, which shows the effectiveness of the multi-modality pre-training.

\begin{figure}[t!]
  \centering
   \includegraphics[width=0.95\linewidth]{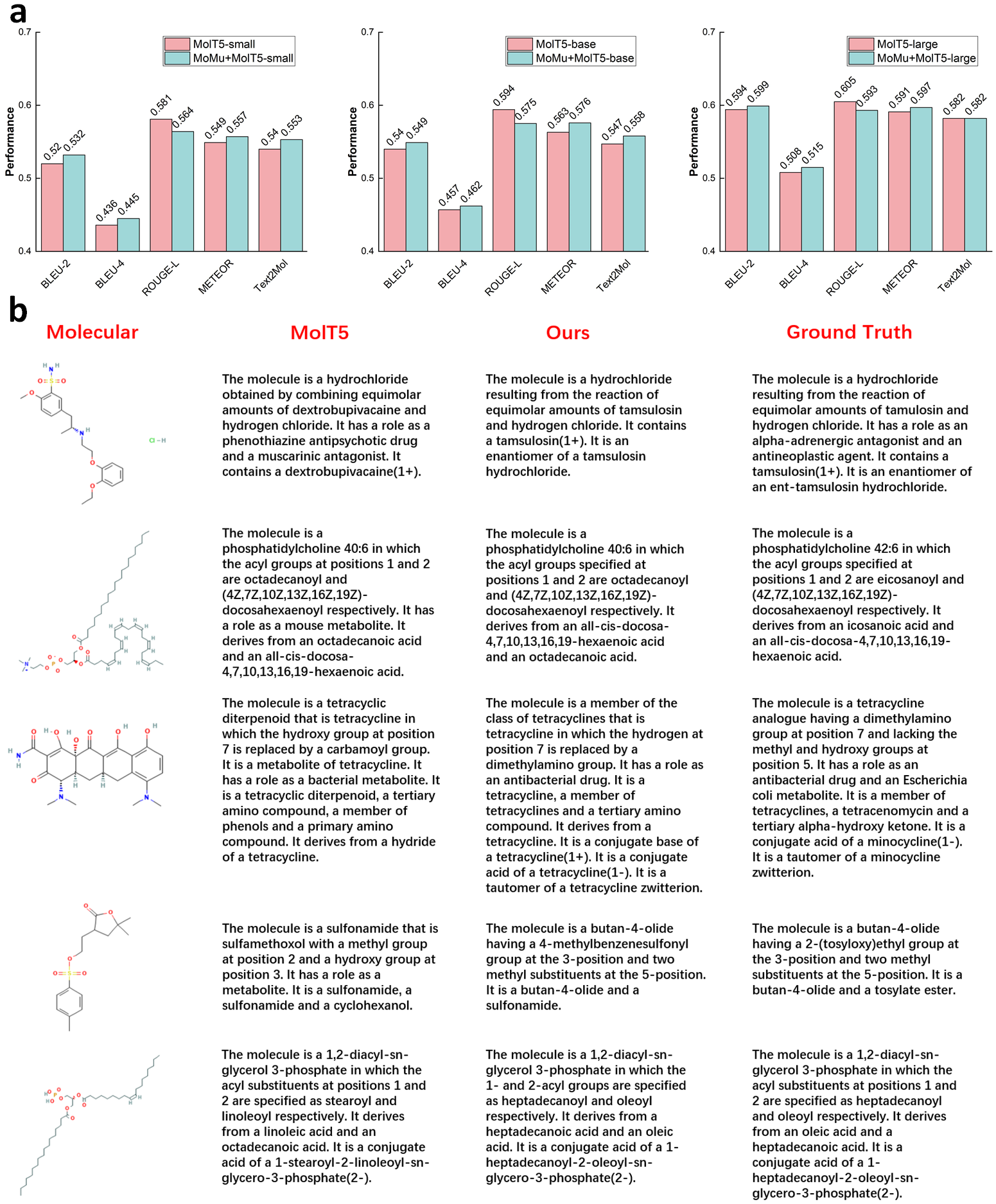}
   \caption{Molecule caption results. (a) Comparison of MolT5 and our MoMu-enhanced MolT5 on the ChEBI-20 dataset. \textit{MolT5} represents the performance with only the MolT5 model, while \textit{MoMu+MolT5} represents the performance after adding GIN-extracted graph features to the input of the MolT5 encoder.  (b) Example captions generated by different models.}
   \label{fig:captionexample}
\end{figure}

\subsection*{Molecule caption}
The molecule captioning task is proposed in~\cite{edwards2022translation}, which aims to generate texts to describe the given molecule. We utilize MolT5 (MolT5-small, MolT5-base, or MolT5-large) \cite{edwards2022translation} as the baseline, which translates the SMILES strings into natural language through the T5-based encoder-decoder transformer architecture. As shown in the Supplementary Figure~3, to better utilize the structural information of the input molecule for translation, we append the graph feature of the molecular graph to the inputs of the MolT5 encoder through a feature mapping module, which is implemented by a multi-layer perceptron. Following~\cite{edwards2022translation}, we evaluate MolT5 and our MoMu-enhanced MolT5 on the ChEBI-20 dataset~\cite{edwards2021text2mol}. Results are shown in Figure~\ref{fig:captionexample}(a). The BLEU, METEOR, and Text2Mol metrics are all improved with the additional graph features, but the ROUGE-L metric is dropped. These results show that incorporating the structural graph information generally leads to smoother and more accurate captions. MoMu-enhanced MolT5 achieves higher scores on the Text2Mol metric, which indicates that the generated captions are more similar to the ground-truth descriptions. This is also evidenced by several examples shown in Figure~\ref{fig:captionexample}(b), where MoMu results in more accurate captions for these molecules with complex structures, e.g., long chains and multiple rings.%, but reduces the recall, which may be due to the loss of some small local structural information in the smoothing process of the graph encoder

%\begin{table}[th]
%  \caption{\textit{MolT5} represents the performance with only the MolT5 model, while \textit{MoMu+MolT5} represents the performance after adding GIN-extracted graph features to the input of the MolT5 encoder. Bigger is better for all the metrics.}%$^\dag$
%  \label{tab:caption}
%  \centering
%  \setlength{\tabcolsep}{1mm}{
%    \begin{tabular}{c|ccccccc}
%    \toprule
%    Model  & BLEU-2 & BLEU-4 & ROUGE-1 & ROUGE-2 & ROUGE-L & METEOR & Text2Mol\\
%    \midrule
%    MolT5-small & 0.520 & 0.436 & {\bf 0.624} & {\bf 0.475} & {\bf 0.581} & 0.549 & 0.540\\
%    MoMu+MolT5-small & {\bf 0.532} & {\bf 0.445} & 0.621 & 0.470 & 0.564 & {\bf 0.557} & {\bf 0.553}\\
%    \midrule
%    MolT5-base & 0.540 & 0.457 & {\bf 0.636} & {\bf 0.489} & {\bf 0.594} & 0.563 & 0.547\\
%    MoMu+MolT5-base & {\bf 0.549} & {\bf 0.462} & 0.629 & 0.482 & 0.575 & {\bf 0.576} & {\bf 0.558}\\
%    \midrule
%    MolT5-large & 0.594 & 0.508 & {\bf 0.650} & {\bf 0.509} & {\bf 0.605} & 0.591 & {\bf 0.582}\\
%    MoMu+MolT5-large & {\bf 0.599} & {\bf 0.515} & 0.643 & 0.500 & 0.593 & {\bf 0.597} & {\bf 0.582}\\
%    \bottomrule
%  \end{tabular}}
%\end{table}

\begin{figure}[t!]
  \centering
   \includegraphics[width=0.98\linewidth]{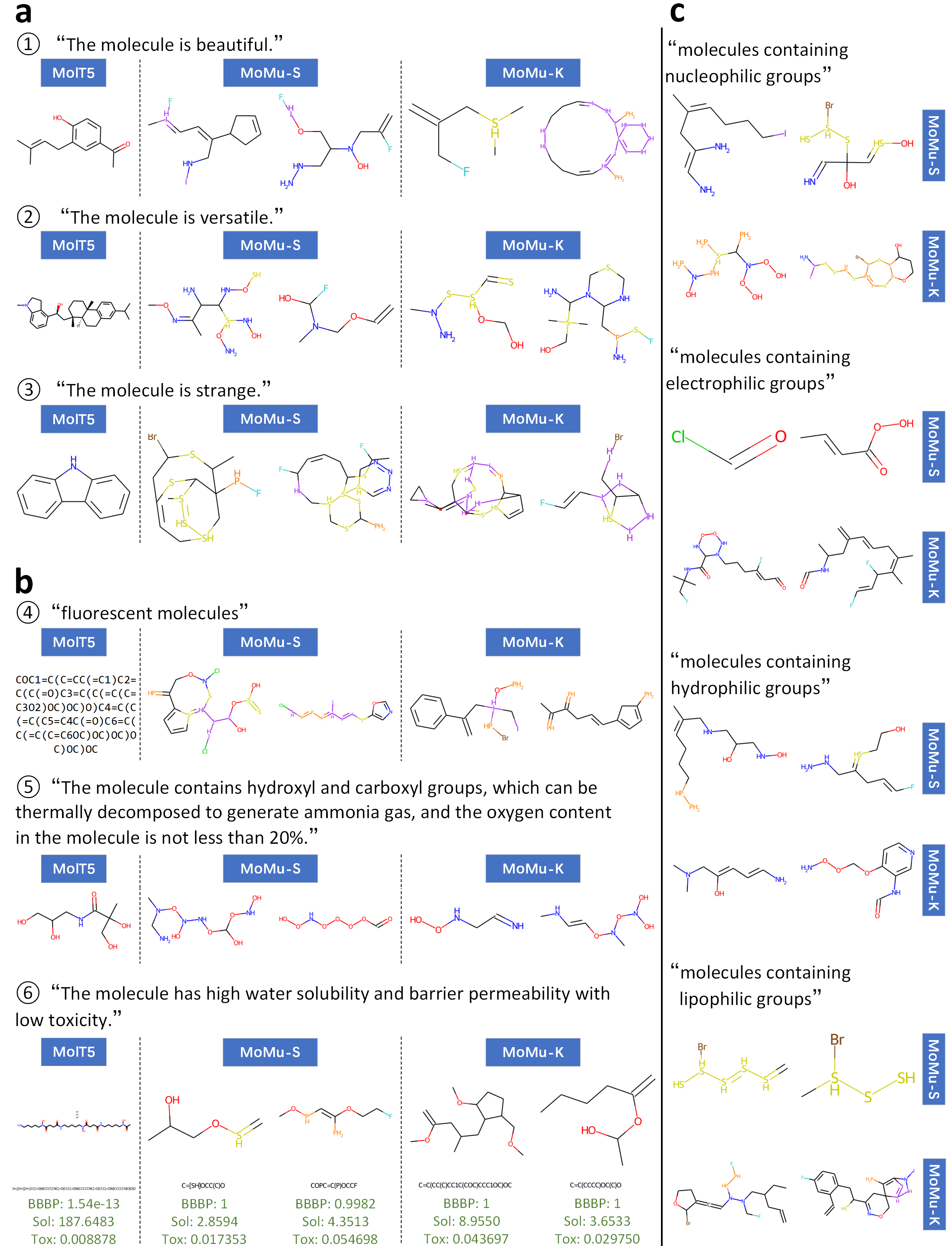}
   \caption{Text-to-graph molecule generation results. (a) Molecules imagined from high-level vague descriptions. (b) Molecules imagined from functional descriptions. (c) Molecules imagined from structural descriptions.}
   \label{fig:generationexample}
\end{figure}
%Left: MolT5, where MolT5-small fails for the first input and MolT5-base fails for the last two inputs; Middle: our zero-shot molecule generation method with MoMu-S; Right: our zero-shot molecule generation method with MoMu-K.

\subsection*{Zero-shot text-to-graph molecule generation}
We propose a new task called zero-shot text-to-graph molecule generation. The goal is to design a cross-modality molecule generator that takes as input the natural language description of the desired conditions and imagines new molecules that match the description. The proposed task is different from the text-to-molecule translation task in~\cite{edwards2022translation}, where the input text is the natural language description of the structure of a ground-truth molecule and the goal is to translate the description into the corresponding molecule. For example, the text ``The molecule is a member of the class of monohydroxy-1,4-benzoquinones that is 2-hydroxy-1,4-benzoquinone carrying an additional methyl substituent at position 5. It is a conjugate acid of a 2-oxido-5-methylquinone.'' exactly describes the molecule ``CC1=CC(=O)C(=O)C=C1O''. Differently, in our zero-shot molecule generation, the natural language text describes the specified properties or conditions of desired new molecules rather than the exact structure of a known molecule. An example of such input description can be ``The molecule has high water solubility and barrier permeability with low toxicity''. Therefore, there may exist different molecules that fit the description. We aim to generate novel molecules that are not present in existing molecule datasets but match the description as closely as possible.%universal

Based on our pre-trained MoMu and a pre-trained molecule generator, we design a zero-shot molecule generation method that does not require additional training data. The architecture of the method is shown in the Supplementary Figure 4, including a MoMu-based multi-modal similarity measuring module and a differentiable molecule generator. We utilize the flow-based molecular generative model, MoFlow~\cite{zang2020moflow}, as the generator, which learns a transformation from a Gaussian distribution in the latent space to the distribution of molecules. Given a random sample $\bm{q}$ from the Gaussian distribution, MoFlow employs the reverse flow to transform $\bm{q}$ to a molecule. To achieve text-to-graph molecule generation, we set $\bm{q}$ as the learnable parameter, and freeze the parameters of both the pre-trained MoMu and MoFlow pre-trained from the ZINC250K dataset~\cite{irwin2012zinc}. We use MoFlow to generate the molecular graph from $\bm{q}$ and then feed it into the graph encoder of MoMu to obtain the graph representation. The input text description is fed into the text encoder of MoMu to obtain the text representation. We optimize $\bm{q}$ by maximizing the cosine similarity between the graph representation and the text representation. The optimized $\bm{q}$ is fed into MoFlow to generate the final molecule. We prohibit the inclusion of formal charges in the resulting molecule. For the detailed algorithm, please refer to the Methods section.%We randomly initialize $\bm{q}$ 20 times to generate 20 different molecules per text input.

%\begin{figure}[ht]
%  \centering
%   \includegraphics[width=0.9\linewidth]{./PretrainFig/GenerationResultsNew2.png}
%   \caption{Example of molecules generated by different methods from three input texts, respectively. Left: MolT5-large; Middle: our zero-shot molecule generation method with MoMu-S; Right: our zero-shot molecule generation method with MoMu-K. For the last query, BBBP, Sol, and Tox indicate the probability of penetrability, the log molecular weight per liter of water, and the average of 12 toxicities, respectively, predicted by the fine-tuned models in~\cite{suresh2021adversarial,gao2022bootstrapping}.}
%   \label{fig:generationexample2}
%\end{figure}

In Figure~\ref{fig:generationexample}(a), we show the generated molecules from three high-level vague descriptions by MolT5 (large)~\cite{edwards2022translation} and our method with MoMu-S and MoMu-K, respectively. Given the input text, MolT5 can only output one molecule. Differently, our method can generate diverse molecules. The results indicate that our method has established an understanding of abstract concepts and, to a certain extent, its own aesthetic. For example, our method tends to regard locally symmetric and stretched molecular graphs as ``beautiful'', view molecular graphs with more different elements and groups as ``versatile'', and treat molecular graphs with many irregular connections, cluttered branches, and irregular terminals as ``strange''. Since MolT5 is trained to translate from a structural description to a specific molecule, it fails to generate explainable molecules from high-level vague descriptions.%MolT5 has three versions with different model sizes, which are denoted by MolT5-small, MolT5-base, and MolT5-large, respectively.any version of For each of the three text inputs, one version of MolT5 fails to generate a valid molecule. D Besides the three molecules with the highest similarities scores, we also show three other results of the 20 molecules that well match the corresponding description from a human perspective.MoMu-S treats molecules with very symmetrical structures and multiple halogen elements as ``strange''. MoMu-K

In Figure~\ref{fig:generationexample}(b), we show the generated molecules from descriptions of molecular functionalities. For the 4-th description, MolT5 fails to generate a valid molecule, while our methods generate molecules with conjugated double bonds or conjugated molecules. For the 5-th description with four conditions, MoMus successfully generate diverse molecules with hydroxyl groups, high oxygen content, and nitrogen to generate ammonia, so three out of four conditions are met. Different from existing AI-based molecule generation methods that are specified for fixed properties, our method adaptively generates molecular candidates based on the input text which can describe any desired one or several conditions. In the 6-th description, we specify three desired properties including high water solubility, high barrier permeability, and low toxicity, which can be evaluated by fine-tuned property prediction models in~\cite{suresh2021adversarial,gao2022bootstrapping}. The molecule generated by MolT5 has low penetration. Our method with both MoMu-S and MoMu-K can generate different molecules that are highly penetrating, have low toxicity, and are positively water-soluble.%For the 5-th description with four conditions, MoMus successfully generate diverse molecules with high oxygen content and hydroxyl groups. Each molecule contains either the carboxyl group or sufficient nitrogen to generate ammonia, so three out of four conditions are met.custom and specified

In Figure~\ref{fig:generationexample}(c), we show the generated molecules from descriptions of molecular structures. For the description of including nucleophilic groups, both MoMu-S and MoMu-K generate diverse molecules with amino groups, hydroxyl groups, or double bond. For the description of including electrophilic groups, both MoMu-S and MoMu-K generate diverse molecules with carbonyl groups, alkyl-like groups, or halogen atoms, even though formal charges are prohibited. For the description of including hydrophilic groups, both MoMu-S and MoMu-K generate molecules containing Hydroxyl, Amino, or Aldehyde with diverse structures. For the description of including lipophilic groups, both MoMu-S and MoMu-K generate molecules containing alkyl-like groups, halogen atoms, or benzene ring with diverse structures.
%Whereas the 20 molecules randomly generated by our method include those with similar functions or properties as described. For example, for the 4-th description, MoMu can successfully generate nucleophilic groups such as OH and NH2.

\begin{figure}[t!]
  \centering
   \includegraphics[width=0.92\linewidth]{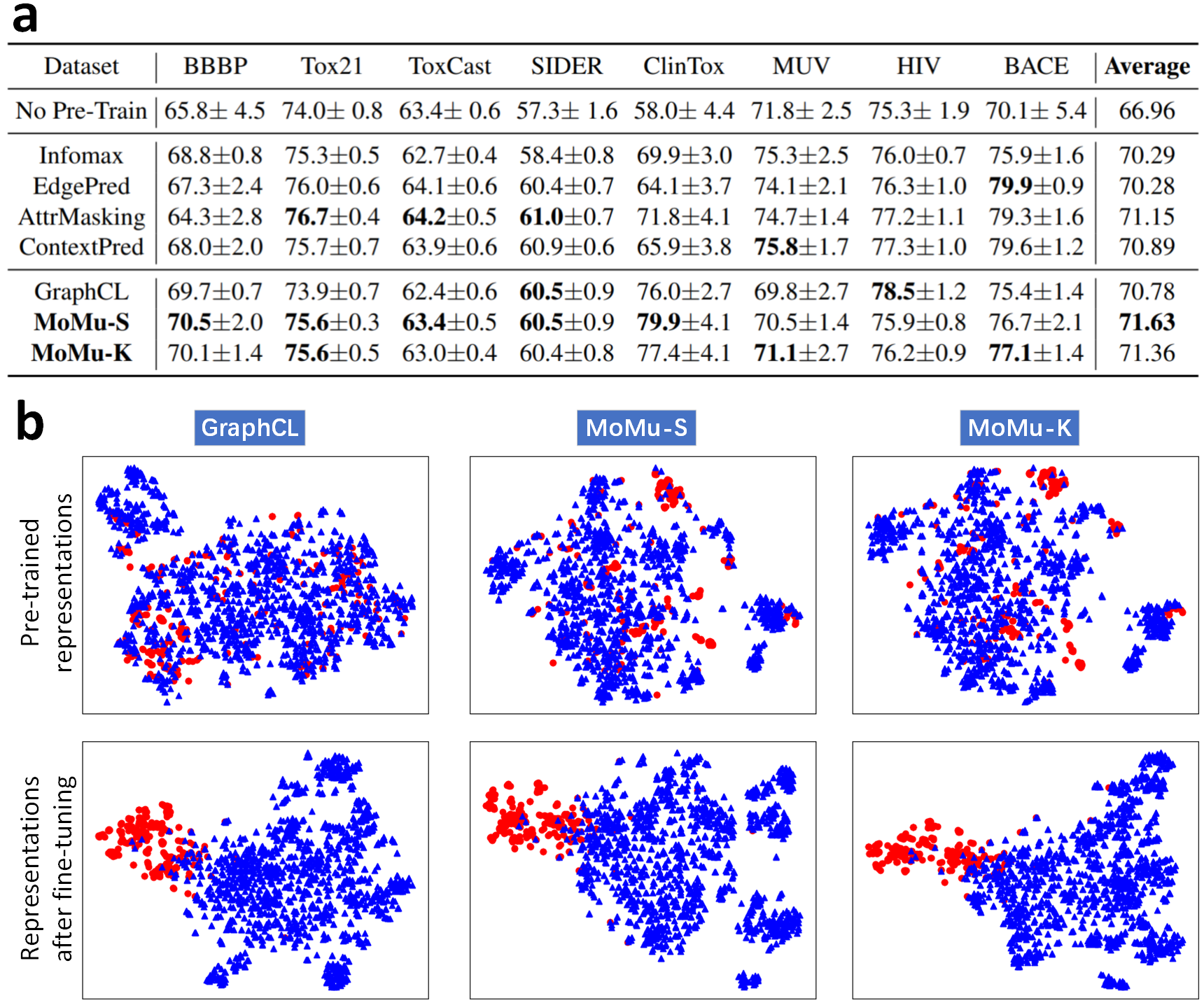}
   \caption{Molecule property prediction results. (a) The performance of molecule property prediction under different pre-training schemes on the eight datasets, where the results of the compared methods are reported in \cite{hu2020pretraining,you2020graph}. (b) The t-SNE visualization of the pretrained representations by different pre-training schemes before and after fine-tuning on the BBBP dataset.}
   \label{fig:predictionresults}
\end{figure}

\subsection*{Molecule property prediction}
Molecular property prediction is a graph-level prediction task that is usually used to evaluate the transferability of pre-trained graph encoders. We perform experiments on eight widely used datasets from MoleculeNet~\cite{wu2018moleculenet}, including BBBP~\cite{martins2012bayesian}, Tox21~\cite{tox21}, ToxCast~\cite{richard2016toxcast}, SIDER~\cite{kuhn2016sider}, ClinTox~\cite{novick2013sweetlead, aact}, MUV~\cite{gardiner2011effectiveness}, HIV~\cite{hiv}, and BACE~\cite{subramanian2016computational}, ranging from predicting bioactivity to toxicity and pharmacokinetics of molecules. Following~\cite{hu2020pretraining, you2020graph}, we split each dataset into training and testing sets according to the scaffold of molecules, which can better reveal the ability of handling out-of-distribution molecules. To apply MoMu, we use the graph encoder in the pre-trained MoMu-S and MuMu-K as the initialization, respectively. We then fine-tune the graph encoder on the training sets of these datasets for predicting molecular properties, respectively. We compare MoMu with different graph-pretraining methods, including Infomax~\cite{velickovic2019deep}, EdgePred~\cite{hamilton2017inductive}, AttrMasking~\cite{hu2020pretraining}, ContextPred~\cite{hu2020pretraining}, and GraphCL~\cite{you2020graph}. All methods use the same GIN encoder.

We conduct experiments 10 times and report the mean and standard deviation of ROC-AUC scores (\%) following \cite{hu2020pretraining}. As shown in Figure \ref{fig:predictionresults}(a), the proposed MoMu-S and MoMu-K outperform the random initialization on most datasets and achieve the best results on 3 datasets. In particular, the graph encoder of MoMu is initialized with weights pre-trained by GraphCL during our pre-training process. Compared with GraphCL, the performance of MoMu drops slightly on the HIV dataset, is comparable on the SIDER dataset, but is better on all the other six datasets. On average, MoMu-S and MoMu-K outperform all other compared methods. This indicates that the multimodal joint pre-training of MoMu captures useful information in weakly-correlated text descriptions for a better understanding of the abstract graph. The graph encoders in MoMu-S and MoMu-K obtain comparable results, which shows that the ability to model SMILES strings can not readily generalized to graphs and our MoMu successfully improves the representational ability of the graph encoder from the text modality.

In Figure \ref{fig:predictionresults}(b), we employ t-SNE~\cite{2008Visualizing} to show the visualizations of the pretrained representations by GraphCL, MoMu-S, and MoMu-K before and after fine-tuning, respectively, on the training data of the BBBP dataset. Compared with the pretrained features by GraphCL, representations learned by MoMu-S and MoMu-K are more scattered and more uniformly distributed. Therefore, after fine-tuning, the representations of molecules with different barrier permeability properties are better separated since the overlapping between their distributions is reduced.

%\begin{figure}[t]
%  \centering
%   \includegraphics[width=0.95\linewidth]{./PretrainFig/RevisedFigure/PredictionResults.png}
%   \caption{The performance of molecule property prediction under different pre-training schemes on the eight datasets, where the results of the compared methods are reported in \cite{hu2020pretraining,you2020graph}.}
%   \label{fig:prediction}
%\end{figure}

\section*{Discussion}
We have presented a molecular multi-modal foundation model, namely MoMu, to bridge molecular graphs and natural language descriptions of molecules. MoMu imitates the human learning process from general knowledge to professional and is pre-trained from our collected weakly correlated data to align graph and text representations in a common space. Extensive experimental results on different downstream tasks such as molecule caption, cross-modality retrieval, and molecular property prediction demonstrate the cross-modality transfer ability and the advantage of joint multi-modal modeling of the proposed MoMu model.%To the best of our knowledge, MoMu is the first deep learning model across language and graph modalities for molecules.

Moreover, we have developed a zero-shot text-to-graph molecule generation method based on MoMu, which learns the generative seeds related to the specified text description for a pre-trained generator via leveraging the model's cross-modal ability. Our method is compatible with any molecule generator that allows gradient back-propagation. The broader the space that a molecular generator can efficiently explore, the more likely our method is to locate the region in which molecules are related to the conditions specified in the text. Since any conditions can be described in the input text, our method allows for convenient custom molecule design. Owing to these advantages, our method may have a significant impact on fields such as drug discovery and materials design that require new molecules with specified properties. %For example, in drug design, different molecules with different properties are needed for different diseases, but it is difficult to collect a large number of molecules with specified properties for each disease. Our text-to-graph generation method shows the potential to generate molecule candidates for the text describing desired conditions for any specific disease, and hence has broad application prospects.

%For generating molecules with specified properties, existing AI-based molecule generation models need to be retrained by collecting data with annotations on those properties. Differently, our method learns the generative seeds related to the specified text description via leveraging the cross-modal ability of MoMu, and then uses a pre-trained general molecule generator to obtain molecules. Simply given a description of the conditions the desired molecule is expected to meet, our method can generate corresponding molecule candidates without any new molecule data or specific retraining.

Besides these advantages and potentials, the proposed molecular multi-modal foundation model also suffers from some limitations. Compared with paired image-text data, the amount of our collected graph-text data for pre-training is much smaller. Even though this problem may be mitigated by initializing from pre-trained unimodal encoders, MoMu may still not be sufficiently trained to fully establish the common space of the molecular graph modality and the natural language modality. In addition, the retrieved text of a molecule may not describe its properties or structure, although the text contains the name of the molecule. Therefore, MoMu can be misled to learn spurious correlations and stereotyped bias against pre-training data. Our zero-shot text-to-graph molecule generation method is limited by the transfer ability of MoMu and the pre-trained molecular generator. It is difficult to generate reliable molecules for molecular properties that do not appear or appear infrequently in the training texts. Since our method can be regarded as learning to sample from the latent distribution of the generator, both the quality of the produced molecules and the chemical space that can be explored highly depend on the pre-trained generator. The MoFlow pretrained on zinc250k used in this paper can only generate molecules with a maximum of 38 atoms, and atoms can only be one of the common 9 elements. This also limits the molecular space that our method can generate.
%MoFlow pretrained on zinc250k used in this paper can only generate molecules with a maximum of 38 atoms and th

In our future work, we intend to improve and better utilize MoMu from the following aspects: (1). Collecting larger-scale molecule data in more modalities (e.g., 3D conformation) and retrieving more related text descriptions, which may enable the training of a more powerful molecular multimodal model. (2). Adopting strongly-correlated paired molecular data and more advanced molecule generators to construct more powerful zero-shot text-to-graph molecule design methods. (3). Developing interpretable tools to reveal the structure of the learned cross-modal common space and how MoMu bridges molecule structures to textural descriptions of molecular properties. (4). Applying the pre-trained MoMu to more downstream tasks and real-world specific cases, e.g., validating the molecules generated by our zero-shot generation method via wet experiments in designing drugs for some special diseases.

\section*{Methods}
\textbf{Data collection.}
PubChem~\cite{pubchem} contains the basic information of over 150 million chemicals. Only simple physical and chemical properties of molecules are annotated in the PubChem database and there are no exact language descriptions available for most molecules. We extract papers in the fields of Medicine, Biology, Chemistry, and Computer Science in the S2orc corpus database. To avoid as much as possible the special characters in the text related to the experimental data, we only retrieve from the abstract, introduction, and conclusion sections in each extracted paper. As shown in the Supplementary Figure 1(a), for each molecule, we first use its name as the query to retrieve sentences including this name. Each retrieved sentence and its neighboring sentences are recorded into a document as a paragraph. If less than two pieces of paragraphs are retrieved by name, we then search by the synonyms of the molecule. When 5000 pieces of paragraphs are retrieved or the size of the document exceeds 500Mb, the retrieval of the molecule is early terminated. Not all 50,000 molecules can be retrieved with corresponding text descriptions. In each retrieved molecule graph-document pair, sentences in the document reveal weakly-correlated semantic information of the corresponding molecule graph.

\textbf{Graph augmentation.}
Each molecular graph needs to undergo two different random augmentations into two augmented molecular graphs that retain similar semantics to the original molecule. We utilize the data augmentations introduced by GraphCL~\cite{you2020graph}. Specifically, we adopt two types of graph augmentations considering priors of molecules, i.e., node dropping and subgraph. Node dropping randomly discards a certain portion of vertices of the original graph. For a molecular graph, missing certain atoms (e.g., some hydrogen atoms in the chemical compound) does not alter its semantic information. Subgraph means sampling a subgraph from the original graph using the random walk. The properties of a molecule have certain similarities to the properties of molecules formed by its subgraphs, e.g., some molecules that contain the same functional group. Therefore, these two augmentations are more suitable for molecular-related tasks, which has been demonstrated in GraphCL~\cite{you2020graph}.

\textbf{Graph encoder.}
Graph Isomorphism Network (GIN) \cite{xu2018powerful} is a provably powerful graph neural network (GNN) under the neighborhood aggregation framework, which has become a general backbone network in the graph domain, especially in molecular graph related tasks. Specifically, we utilize atom number and atom chirality as node features to feed into several GIN convolutional layers involving bond type and attribute features. Finally, we read out the feature of the entire molecular graph by performing average pooling of all node features.

\textbf{Text encoder.}
The Bert model \cite{devlin2018bert} is widely used as the feature extractor in natural language processing. One BERT variation, Sci-BERT \cite{beltagy2019scibert}, is pre-trained using a total of 1.14 million scientific papers in the domains of biomedicine (82\%) and computer science (12\%) and is better suited for processing texts that describe molecular properties. KV-PLM~\cite{zeng2022deep} further jointly models natural languages and SMILES of molecules with BERT. We employ Sci-BERT and KV-PLM as the initialized text encoder to extract the sentence feature sequences, respectively, followed by average pooling to obtain the sentence features.

\textbf{Cross-modal Supervision.}
For the graph modal, a mini-batch of $N$ molecular graphs $\{\bm{\mathcal{G}}_1, \ldots, \bm{\mathcal{G}}_N\}$ are processed by two different augmentations, resulting in $2N$ augmented graphs. Subsequently, we feed these graphs into the graph encoder to obtain their representation vectors $\{\bm{z}_1^G, \bm{\tilde z}_1^G, \ldots, \bm{z}_N^G, \bm{\tilde z}_N^G\}$, where $\bm{z}_i^G$ and $\bm{\tilde z}_i^G$ denote the representations of two augmented versions from the $i$-th graph $\bm{\mathcal{G}}_i$. Meanwhile, the representations obtained by passing the two different sentences describing $\bm{\mathcal{G}}_i$ through the text encoder are denoted as $\bm{z}_i^T$, $\bm{\tilde z}_i^T$. There are two different sentences corresponding to each molecular graph in a mini-batch, resulting in $2N$ text representations $\{\bm{z}_1^T, \bm{\tilde z}_1^T, \ldots, \bm{z}_N^T, \bm{\tilde z}_N^T\}$. Therefore, for the $i$-th graph $\bm{\mathcal{G}}_i$, the total multi-view loss includes four contrastive losses between four paris of representations from multi-modals, i.e., $(\bm{z}_i^G, \bm{z}_i^T)$, $(\bm{\tilde z}_i^G, \bm{z}_i^T)$, $(\bm{z}_i^G, \bm{\tilde z}_i^T)$ and $(\bm{\tilde z}_i^G, \bm{\tilde z}_i^T)$. For the sake of simplicity, we only denote the contrastive loss for $(\bm{z}_i^G, \bm{z}_i^T)$ as:
\begin{equation}
    \ell_i^{(\bm{z}_i^G, \bm{z}_i^T)} = -\log \frac{\exp{(\emph{sim}(\bm{z}_i^G,\bm{z}_i^T)/\tau)}}{\sum_{j=1}^{N}\exp{(\emph{sim}(\bm{z}_i^G,\bm{z}_j^T)/\tau})},
\end{equation}
where $\tau$ is the temperature parameter and $\emph{sim}(\bm{z}_i^G,\bm{z}_i^T)$ first passes $\bm{z}_i^G$ and $\bm{z}_i^T$ into two separate projection heads to project them into the same dimension, and then calculates the cosine similarity between the projected vectors. The other three cross-modal contrastive losses have the same form.

\textbf{Graph-modal Self-Supervision.}
To enhance the representational power of the graph encoder further, we utilize contrastive learning within the graph modality. Specifically, we pull in the features of positive pairs while pushing those of negative pairs away by minimizing the normalized temperature-scaled cross-entropy loss \cite{wu2018unsupervised,oord2018representation,you2020graph}, where positive pairs are two augmentations of the same molecular graph and negative pairs come from different molecular graphs. Based on the previous definitions, we derive the graph-modal contrastive loss for the $i$-th graph as:
\begin{equation}
    \ell_i^{(\bm{z}_i^G, \bm{\tilde z}_i^G)} = -\log \frac{\exp{(\emph{sim}(\bm{z}_i^G,\bm{\tilde z}_i^G)/\tau)}}{\sum_{j=1}^{N}\exp{(\emph{sim}(\bm{z}_i^G,\bm{\tilde z}_j^G)/\tau})},
\end{equation}
where $\tau$ is the temperature parameter and the final loss is computed across all samples in the mini-batch.

\textbf{Implementation details.} Following GraphCL \cite{you2020graph}, we use the GIN with 5 layers and a 300-dimensional hidden size as the graph encoder. We select the base BERT \footnote{\url{https://huggingface.co/bert-base-uncased.}}, whose hidden size is 768, for the text encoder. Graph features and sentence features are projected into the same feature space using two multi-layer perceptrons, each of whose output dimension is 256. Prior to the pre-training procedure, we initialize the GIN model using the GraphCL checkpoint \footnote{\url{https://github.com/Shen-Lab/GraphCL/tree/master/transferLearning_MoleculeNet_PPI/chem/models_graphcl}} and the BERT model using the checkpoints of Sci-BERT \cite{beltagy2019scibert} or KV-PLM \cite{zeng2022deep}. Then we pre-train the models using the dataset we have gathered, which includes pairs of molecular graphs and texts. For two graph augmentations, the node dropping ratio is 10\% and the size of the sampled subgraph is 80\% of the original graph. The input data for the text modal is two sentences randomly selected from the document of the corresponding graph data. We employ the AdamW optimizer with a learning rate of 0.0001 and a weight decay of 1e-5 for 300 epochs to pre-train our models. $\tau$ is set to 0.1 and the batch size is set to 256. The entire pre-training process is implemented with PyTorch \cite{pytorch} and conducted on eight NVIDIA Tesla V100 PCIe 32GB GPUs.

\textbf{Formalization of zero-shot molecule generation.} As shown in the Supplementary Figure~3, our zero-shot text-to-graph molecule generation method consists of the MoMu-based similarity-measuring module and the MoFlow-based molecule generator. MoFlow defines a parameterized invertible mapping flow from a Gaussian distribution to the distribution of molecues. A molecule graph $\bm{\mathcal{G}}$ is a pair of an atom matrix $\bm{V} \in \mathbb{R}^{N \times C_a}$ and a bond matrix $\bm{E} \in \mathbb{R}^{N \times N \times C_b}$, where $N$ is the number of atoms in the molecule, $C_a$ and $C_b$ are the number of atom types and bond types. $\bm{V}_{n,c}=1$ if the $n$-th atom belongs to the $c$-th atom type and $\bm{V}_{n,c}=0$ otherwise. $\bm{E}_{n,n',c'}=1$ if the bond between the $n$-th atom and the $n'$-th atom belongs to the $c'$-th bond type and $\bm{E}_{n,n',c'}=0$ otherwise. MoFlow contains a graph conditional flow $\bm{q}_v = f_c(\bm{V}|\bm{E})$ for encoding the atom matrix $\bm{V}$ given the bond matrix $\bm{E}$ into a latent variable $\bm{q}_v$ and a gflow $\bm{q}_e = f_g(\bm{E})$ for encoding the bond matrix $\bm{E}$ into a latent variable $\bm{q}_e$. $f_c$ and $f_g$ are implemented by graph coupling layer-based graph convolution neural networks. The concatenation $\bm{q} = [\bm{q}_v; \bm{q}_e]$ of $\bm{q}_v$ and $\bm{q}_e$ follows a Gaussian distribution $P(\bm{q})$. Once MoFlow is trained, a variable $\bm{q}$ can be sampled from $P(\bm{q})$ and decomposed into two parts $\bm{q}_v$ and $\bm{q}_e$, which are fed into the reverse graph conditional flow $f_c^{-1}$ and the reverse gflow $f_g^{-1}$ to obtain the probability matrices:
\begin{equation}
\hat{\bm{E}} = f_g^{-1}(\bm{q}_e) \in \mathbb{R}^{N \times N \times C_b},
\end{equation}%, \bm{E} = GN(argmax(\hat{\bm{E}}))
\begin{equation}
\hat{\bm{V}} = f_c^{-1}(\bm{q}_v|GN(\hat{\bm{E}})) \in \mathbb{R}^{N \times C_a},
\end{equation}
where $\hat{\bm{E}}_{n,n',c'}$ is the predicted probability that the bond between the $n$-th atom and the $n'$-th atom belongs to the $c'$-th bond type, and $\hat{\bm{V}}_{n,c}$ is the probability that $n$-th atom belongs to the $c$-th atom type. $\bm{V}$ and $\bm{E}$ can be obtained from $\hat{\bm{V}}$ and $\hat{\bm{E}}$ by performing the argmax operation to the last dimension, respectively. $GN$ is the graph normalization layer in~\cite{zang2020moflow}. By sampling different $\bm{q}$ from $P(\bm{q})$, MoFlow can generate different novel and valid molecules.

Our zero-shot text-to-graph molecule generation method takes a text description $\bm{x}^T$ as input. The only learnable parameter in our method is $\bm{q}$ which is initialized by randomly sampling from $P(\bm{q})$. All parameters of the pre-trained MoMu and MoFlow are freezed. The input $\bm{x}^T$ is fed into the text encoder of MoMu to obtain the text representation $\bm{z}^T$. $\bm{q}$ is fed into MoFlow to obtain $\hat{\bm{V}}$ and $\hat{\bm{E}}$. To make all operations differentiable and thus allow gradient backpropagation, we fed $\hat{\bm{V}}$ instead of $\bm{V}$ into the graph encoder of MoMu to obtain the graph representation $\bm{z}^G$. The trained graph encoder, GIN, contains embeddings for all atom and bond types. Originally, $\bm{V}$ is used as the indicator to select the corresponding embeddings with respect to the atom types in the first layer. When $\hat{\bm{V}}$ is used, for each atom, the representation is actually the weighted sum of all atom embeddings. The probabilities among atom types per node in $\hat{\bm{V}}$ serve as attention scores. The loss function is the cosine similarity between the projected $\bm{z}^T$ and $\bm{z}^G$:
\begin{equation}
\ell_q = -\emph{sim}(\bm{z}^G,\bm{z}^T)/\tau,
\end{equation}
where $\emph{sim}(\bm{z}^G,\bm{z}^T)$ is the module for calculating the similarity between the projected representations in MoMu. $\bm{q}$ can be updated through gradient backpropagation with respect to $\ell_q$. We use the Adam optimizer for updating. After repeating the update for a maximize of $500$ iterations, we obtain the optimized $\bm{q}^*$, which is then fed into MoFlow to obtain $\hat{\bm{V}}$ and $\hat{\bm{E}}$. Finally, the molecule graph ($\bm{\mathcal{G}} = {\bm{V}, \bm{E}}$) is obtained by performing the argmax operations to the last dimension of $\hat{\bm{V}}$ and $\hat{\bm{E}}$. The pipeline is summarized in Algorithm~\ref{alg:zmg}.%and $\hat{\bm{E}}$ and $\bm{E}$ and $\bm{E}$ are and bond and $\hat{\bm{E}}$ are or bond or bond or bond or $\hat{\bm{E}}$

\begin{algorithm}[t!]
	\vskip 0.in
	\begin{algorithmic}
	\STATE {\bfseries Input:} The pretrained graph encoder $f^G$, text encoder $f^T$, and similarity calculation module $\emph{sim}$ of MoMu; The pretrained graph conditional flow network $f_c$, the gflow network $f_g$, and Gaussian distribution $P(\bm{q})$ of MoFlow; The input text description $\bm{x}^T$; Hyper-parameters: the learning rate $l_r$, the maximum number of iteration $T_m$, the temperature $\tau$.\\
		\STATE {\bf Initialize} The learnable parameter $\bm{q}$ randomly sampled from $P(\bm{q})$.
        \STATE Obtain the text representation $\bm{z}^T = f^T(\bm{x}^T)$
        \FOR{iteration = $1, \cdots, T_m$}
        \STATE Separate $\bm{q}$ into $\bm{q}_v$ and $\bm{q}_e$
        \STATE $\hat{\bm{E}} = f_g^{-1}(\bm{q}_e)$
        \STATE $\bm{E} =argmax(\hat{\bm{E}})$
        \STATE $\hat{\bm{V}} = f_c^{-1}(\bm{q}_v|GN(\hat{\bm{E}}))$
        \STATE Obtain the graph representation $\bm{z}^G = f^G(\hat{\bm{V}}, {\bm{E}})$
        \STATE Calculate the loss $\ell_q = -\emph{sim}(\bm{z}^G,\bm{z}^T)/\tau$
        \STATE Update $\bm{q}$ by the Adam optimizer with $l_r$
        \ENDFOR
        \STATE Separate $\bm{q}$ into $\bm{q}_v$ and $\bm{q}_e$
        \STATE $\hat{\bm{E}} = f_g^{-1}(\bm{q}_e)$
        \STATE $\bm{E} = argmax(\hat{\bm{E}})$
        \STATE $\hat{\bm{V}} = f_c^{-1}(\bm{q}_v|GN(\hat{\bm{E}}))$
        \STATE $\bm{V} = argmax(\hat{\bm{V}})$
        \STATE {\bf Return} the generated molecule graph $\bm{\mathcal{G}} = {\bm{V}, \bm{E}}$		
	\end{algorithmic}
	\vskip -0.in
	\caption{Zero-shot text-to-graph molecule generation.}
	\label{alg:zmg}
\end{algorithm}

\section*{Data availability}
For collecting the pretraining data, the PubChem dataset is available at \url{https://pubchem.ncbi.nlm.nih.gov/}, and the S2orc dataset is available at \url{https://github.com/allenai/s2orc}. Our collected dataset consists of two folders holding molecular graphs and texts, respectively. The dataset can be downloaded in \url{https://pan.baidu.com/s/1aHJoYTTZWDHPCcRuu9I7Fg}. For cross-modality retrieval, the PCdes dataset is available at \url{https://github.com/thunlp/KV-PLM}. For molecule caption, the ChEBI-20 dataset is available at \url{https://github.com/cnedwards/text2mol}. For text-to-molecule generation, the pre-trained MoFlow is available at \url{https://github.com/calvin-zcx/moflow}. For molecule property prediction, the eight datasets from MoleculeNet are available at \url{https://github.com/deepchem/deepchem/tree/master/datasets} and the processed datasets are available at: \url{http://snap.stanford.edu/gnn-pretrain/data/chem_dataset.zip}.

\section*{Code availability}
The code for training the proposed molecular multimodal foundation model and employing the trained model in molecule caption, zero-shot text-to-graph generation, and molecule property prediction will be available at \url{https://github.com/BingSu12/MoMu}. The code for data collection and cross-modality retrieval will be available at \url{https://github.com/yangzhao1230/GraphTextRetrieval}.

{\small
\bibliographystyle{unsrt}
\bibliography{ref_pretrain}

\begin{thebibliography}{10}

\bibitem{hajduk2007decade}
Philip~J Hajduk and Jonathan Greer.
\newblock A decade of fragment-based drug design: strategic advances and
  lessons learned.
\newblock {\em Nature reviews Drug discovery}, 6(3):211--219, 2007.

\bibitem{clark2009design}
Matthew~A Clark, Raksha~A Acharya, Christopher~C Arico-Muendel, Svetlana~L
  Belyanskaya, Dennis~R Benjamin, Neil~R Carlson, Paolo~A Centrella, Cynthia~H
  Chiu, Steffen~P Creaser, John~W Cuozzo, et~al.
\newblock Design, synthesis and selection of dna-encoded small-molecule
  libraries.
\newblock {\em Nature chemical biology}, 5(9):647--654, 2009.

\bibitem{rodrigues2016counting}
Tiago Rodrigues, Daniel Reker, Petra Schneider, and Gisbert Schneider.
\newblock Counting on natural products for drug design.
\newblock {\em Nature chemistry}, 8(6):531--541, 2016.

\bibitem{lecun2015deep}
Yann LeCun, Yoshua Bengio, and Geoffrey Hinton.
\newblock Deep learning.
\newblock {\em nature}, 521(7553):436--444, 2015.

\bibitem{lu2018accelerated}
Shuaihua Lu, Qionghua Zhou, Yixin Ouyang, Yilv Guo, Qiang Li, and Jinlan Wang.
\newblock Accelerated discovery of stable lead-free hybrid organic-inorganic
  perovskites via machine learning.
\newblock {\em Nature communications}, 9(1):1--8, 2018.

\bibitem{jin2020multi}
Wengong Jin, Regina Barzilay, and Tommi Jaakkola.
\newblock Multi-objective molecule generation using interpretable
  substructures.
\newblock In {\em International conference on machine learning}, pages
  4849--4859. PMLR, 2020.

\bibitem{walters2020applications}
W~Patrick Walters and Regina Barzilay.
\newblock Applications of deep learning in molecule generation and molecular
  property prediction.
\newblock {\em Accounts of chemical research}, 54(2):263--270, 2020.

\bibitem{kotsias2020direct}
Panagiotis-Christos Kotsias, Josep Ar{\'u}s-Pous, Hongming Chen, Ola Engkvist,
  Christian Tyrchan, and Esben~Jannik Bjerrum.
\newblock Direct steering of de novo molecular generation with descriptor
  conditional recurrent neural networks.
\newblock {\em Nature Machine Intelligence}, 2(5):254--265, 2020.

\bibitem{mahmood2021masked}
Omar Mahmood, Elman Mansimov, Richard Bonneau, and Kyunghyun Cho.
\newblock Masked graph modeling for molecule generation.
\newblock {\em Nature communications}, 12(1):1--12, 2021.

\bibitem{wang2019smiles}
Sheng Wang, Yuzhi Guo, Yuhong Wang, Hongmao Sun, and Junzhou Huang.
\newblock Smiles-bert: large scale unsupervised pre-training for molecular
  property prediction.
\newblock In {\em Proceedings of the 10th ACM international conference on
  bioinformatics, computational biology and health informatics}, pages
  429--436, 2019.

\bibitem{chithrananda2020chemberta}
Seyone Chithrananda, Gabriel Grand, and Bharath Ramsundar.
\newblock Chemberta: large-scale self-supervised pretraining for molecular
  property prediction.
\newblock {\em arXiv preprint arXiv:2010.09885}, 2020.

\bibitem{beltagy2019scibert}
Iz~Beltagy, Kyle Lo, and Arman Cohan.
\newblock Scibert: A pretrained language model for scientific text.
\newblock pages 3615--3620, 2019.

\bibitem{li2019biomedical}
Diya Li, Lifu Huang, Heng Ji, and Jiawei Han.
\newblock Biomedical event extraction based on knowledge-driven tree-lstm.
\newblock In {\em Proceedings of the 2019 Conference of the North American
  Chapter of the Association for Computational Linguistics: Human Language
  Technologies, Volume 1 (Long and Short Papers)}, pages 1421--1430, 2019.

\bibitem{lee2020biobert}
Jinhyuk Lee, Wonjin Yoon, Sungdong Kim, Donghyeon Kim, Sunkyu Kim, Chan~Ho So,
  and Jaewoo Kang.
\newblock Biobert: a pre-trained biomedical language representation model for
  biomedical text mining.
\newblock {\em Bioinformatics}, 36(4):1234--1240, 2020.

\bibitem{kusner2017grammar}
Matt~J Kusner, Brooks Paige, and Jos{\'e}~Miguel Hern{\'a}ndez-Lobato.
\newblock Grammar variational autoencoder.
\newblock In {\em International conference on machine learning}, pages
  1945--1954. PMLR, 2017.

\bibitem{dai2018syntax}
Hanjun Dai, Yingtao Tian, Bo~Dai, Steven Skiena, and Le~Song.
\newblock Syntax-directed variational autoencoder for structured data.
\newblock In {\em International Conference on Learning Representations}, 2018.

\bibitem{segler2018generating}
Marwin~HS Segler, Thierry Kogej, Christian Tyrchan, and Mark~P Waller.
\newblock Generating focused molecule libraries for drug discovery with
  recurrent neural networks.
\newblock {\em ACS central science}, 4(1):120--131, 2018.

\bibitem{gomez2018automatic}
Rafael G{\'o}mez-Bombarelli, Jennifer~N Wei, David Duvenaud, Jos{\'e}~Miguel
  Hern{\'a}ndez-Lobato, Benjam{\'\i}n S{\'a}nchez-Lengeling, Dennis Sheberla,
  Jorge Aguilera-Iparraguirre, Timothy~D Hirzel, Ryan~P Adams, and Al{\'a}n
  Aspuru-Guzik.
\newblock Automatic chemical design using a data-driven continuous
  representation of molecules.
\newblock {\em ACS central science}, 4(2):268--276, 2018.

\bibitem{honda2019smiles}
Shion Honda, Shoi Shi, and Hiroki~R Ueda.
\newblock Smiles transformer: Pre-trained molecular fingerprint for low data
  drug discovery.
\newblock {\em arXiv preprint arXiv:1911.04738}, 2019.

\bibitem{popova2019molecularrnn}
Mariya Popova, Mykhailo Shvets, Junier Oliva, and Olexandr Isayev.
\newblock Molecularrnn: Generating realistic molecular graphs with optimized
  properties.
\newblock {\em arXiv preprint arXiv:1905.13372}, 2019.

\bibitem{wang2021multi}
Jike Wang, Chang-Yu Hsieh, Mingyang Wang, Xiaorui Wang, Zhenxing Wu, Dejun
  Jiang, Benben Liao, Xujun Zhang, Bo~Yang, Qiaojun He, et~al.
\newblock Multi-constraint molecular generation based on conditional
  transformer, knowledge distillation and reinforcement learning.
\newblock {\em Nature Machine Intelligence}, 3(10):914--922, 2021.

\bibitem{guo2022improving}
Jeff Guo, Vendy Fialkov{\'a}, Juan~Diego Arango, Christian Margreitter,
  Jon~Paul Janet, Kostas Papadopoulos, Ola Engkvist, and Atanas Patronov.
\newblock Improving de novo molecular design with curriculum learning.
\newblock {\em Nature Machine Intelligence}, pages 1--9, 2022.

\bibitem{flam2022language}
Daniel Flam-Shepherd, Kevin Zhu, and Al{\'a}n Aspuru-Guzik.
\newblock Language models can learn complex molecular distributions.
\newblock {\em Nature Communications}, 13(1):1--10, 2022.

\bibitem{hoffman2022optimizing}
Samuel~C Hoffman, Vijil Chenthamarakshan, Kahini Wadhawan, Pin-Yu Chen, and
  Payel Das.
\newblock Optimizing molecules using efficient queries from property
  evaluations.
\newblock {\em Nature Machine Intelligence}, 4(1):21--31, 2022.

\bibitem{hy2018predicting}
Truong~Son Hy, Shubhendu Trivedi, Horace Pan, Brandon~M Anderson, and Risi
  Kondor.
\newblock Predicting molecular properties with covariant compositional
  networks.
\newblock {\em The Journal of chemical physics}, 148(24):241745, 2018.

\bibitem{coley2019graph}
Connor~W Coley, Wengong Jin, Luke Rogers, Timothy~F Jamison, Tommi~S Jaakkola,
  William~H Green, Regina Barzilay, and Klavs~F Jensen.
\newblock A graph-convolutional neural network model for the prediction of
  chemical reactivity.
\newblock {\em Chemical science}, 10(2):370--377, 2019.

\bibitem{zaidi2022pre}
Sheheryar Zaidi, Michael Schaarschmidt, James Martens, Hyunjik Kim, Yee~Whye
  Teh, Alvaro Sanchez-Gonzalez, Peter Battaglia, Razvan Pascanu, and Jonathan
  Godwin.
\newblock Pre-training via denoising for molecular property prediction.
\newblock {\em arXiv preprint arXiv:2206.00133}, 2022.

\bibitem{li2022glam}
Yuquan Li, Chang-Yu Hsieh, Ruiqiang Lu, Xiaoqing Gong, Xiaorui Wang, Shuo Liu,
  Yanan Tian, Dejun Jiang, Jiaxian Yan, Qifeng Bai, et~al.
\newblock Glam: An adaptive graph learning method for automated molecular
  interactions and properties predictions.
\newblock {\em Nature Machine Intelligence}, 2022.

\bibitem{wang2021molclr}
Yuyang Wang, Jianren Wang, Zhonglin Cao, and Amir~Barati Farimani.
\newblock Molecular contrastive learning of representations via graph neural
  networks.
\newblock {\em Nature Machine Intelligence}, 2022.

\bibitem{wang2021chemical}
Hongwei Wang, Weijiang Li, Xiaomeng Jin, Kyunghyun Cho, Heng Ji, Jiawei Han,
  and Martin Burke.
\newblock Chemical-reaction-aware molecule representation learning.
\newblock In {\em International Conference on Learning Representations}, 2021.

\bibitem{chen2021algebraic}
Dong Chen, Kaifu Gao, Duc~Duy Nguyen, Xin Chen, Yi~Jiang, Guo-Wei Wei, and Feng
  Pan.
\newblock Algebraic graph-assisted bidirectional transformers for molecular
  property prediction.
\newblock {\em Nature Communications}, 12(1):1--9, 2021.

\bibitem{gilmer2017neural}
Justin Gilmer, Samuel~S Schoenholz, Patrick~F Riley, Oriol Vinyals, and
  George~E Dahl.
\newblock Neural message passing for quantum chemistry.
\newblock In {\em International conference on machine learning}, pages
  1263--1272. PMLR, 2017.

\bibitem{jin2018junction}
Wengong Jin, Regina Barzilay, and Tommi Jaakkola.
\newblock Junction tree variational autoencoder for molecular graph generation.
\newblock In {\em International conference on machine learning}, pages
  2323--2332. PMLR, 2018.

\bibitem{shi2019graphaf}
Chence Shi, Minkai Xu, Zhaocheng Zhu, Weinan Zhang, Ming Zhang, and Jian Tang.
\newblock Graphaf: a flow-based autoregressive model for molecular graph
  generation.
\newblock In {\em International Conference on Learning Representations}, 2019.

\bibitem{zang2020moflow}
Chengxi Zang and Fei Wang.
\newblock Moflow: an invertible flow model for generating molecular graphs.
\newblock In {\em Proceedings of the 26th ACM SIGKDD International Conference
  on Knowledge Discovery \& Data Mining}, pages 617--626, 2020.

\bibitem{ma2021gf}
Changsheng Ma and Xiangliang Zhang.
\newblock Gf-vae: A flow-based variational autoencoder for molecule generation.
\newblock In {\em Proceedings of the 30th ACM International Conference on
  Information \& Knowledge Management}, pages 1181--1190, 2021.

\bibitem{luo2021graphdf}
Youzhi Luo, Keqiang Yan, and Shuiwang Ji.
\newblock Graphdf: A discrete flow model for molecular graph generation.
\newblock In {\em International Conference on Machine Learning}, pages
  7192--7203. PMLR, 2021.

\bibitem{chen2021deep}
Ziqi Chen, Martin~Renqiang Min, Srinivasan Parthasarathy, and Xia Ning.
\newblock A deep generative model for molecule optimization via one fragment
  modification.
\newblock {\em Nature Machine Intelligence}, 3(12):1040--1049, 2021.

\bibitem{lee2022mgcvae}
Myeonghun Lee and Kyoungmin Min.
\newblock Mgcvae: Multi-objective inverse design via molecular graph
  conditional variational autoencoder.
\newblock {\em Journal of Chemical Information and Modeling}, 2022.

\bibitem{zeng2022deep}
Zheni Zeng, Yuan Yao, Zhiyuan Liu, and Maosong Sun.
\newblock A deep-learning system bridging molecule structure and biomedical
  text with comprehension comparable to human professionals.
\newblock {\em Nature communications}, 13(1):1--11, 2022.

\bibitem{s2ocr}
Kyle Lo, Lucy~Lu Wang, Mark Neumann, Rodney~Michael Kinney, and Daniel~S. Weld.
\newblock S2orc: The semantic scholar open research corpus.
\newblock In {\em ACL}, 2020.

\bibitem{radford2021learning}
Alec Radford, Jong~Wook Kim, Chris Hallacy, Aditya Ramesh, Gabriel Goh,
  Sandhini Agarwal, Girish Sastry, Amanda Askell, Pamela Mishkin, Jack Clark,
  et~al.
\newblock Learning transferable visual models from natural language
  supervision.
\newblock In {\em International Conference on Machine Learning}, pages
  8748--8763. PMLR, 2021.

\bibitem{jia2021scaling}
Chao Jia, Yinfei Yang, Ye~Xia, Yi-Ting Chen, Zarana Parekh, Hieu Pham, Quoc Le,
  Yun-Hsuan Sung, Zhen Li, and Tom Duerig.
\newblock Scaling up visual and vision-language representation learning with
  noisy text supervision.
\newblock In {\em International Conference on Machine Learning}, pages
  4904--4916. PMLR, 2021.

\bibitem{li2020oscar}
Xiujun Li, Xi~Yin, Chunyuan Li, Pengchuan Zhang, Xiaowei Hu, Lei Zhang, Lijuan
  Wang, Houdong Hu, Li~Dong, Furu Wei, et~al.
\newblock Oscar: Object-semantics aligned pre-training for vision-language
  tasks.
\newblock In {\em European Conference on Computer Vision}, pages 121--137.
  Springer, 2020.

\bibitem{fei2022towards}
Nanyi Fei, Zhiwu Lu, Yizhao Gao, Guoxing Yang, Yuqi Huo, Jingyuan Wen, Haoyu
  Lu, Ruihua Song, Xin Gao, Tao Xiang, et~al.
\newblock Towards artificial general intelligence via a multimodal foundation
  model.
\newblock {\em Nature Communications}, 13(1):1--13, 2022.

\bibitem{you2020graph}
Yuning You, Tianlong Chen, Yongduo Sui, Ting Chen, Zhangyang Wang, and Yang
  Shen.
\newblock Graph contrastive learning with augmentations.
\newblock {\em Advances in Neural Information Processing Systems},
  33:5812--5823, 2020.

\bibitem{pubchem}
Yanli Wang, Jewen Xiao, Tugba~O Suzek, Jian Zhang, Jiyao Wang, and Stephen~H
  Bryant.
\newblock Pubchem: a public information system for analyzing bioactivities of
  small molecules.
\newblock {\em Nucleic acids research}, 37(suppl\_2):W623--W633, 2009.

\bibitem{ogb}
Weihua Hu, Matthias Fey, Marinka Zitnik, Yuxiao Dong, Hongyu Ren, Bowen Liu,
  Michele Catasta, and Jure Leskovec.
\newblock Open graph benchmark: Datasets for machine learning on graphs.
\newblock {\em Advances in neural information processing systems},
  33:22118--22133, 2020.

\bibitem{xu2018powerful}
Keyulu Xu, Weihua Hu, Jure Leskovec, and Stefanie Jegelka.
\newblock How powerful are graph neural networks?
\newblock In {\em International Conference on Learning Representations}, 2018.

\bibitem{devlin2018bert}
Jacob Devlin Ming-Wei~Chang Kenton and Lee~Kristina Toutanova.
\newblock Bert: Pre-training of deep bidirectional transformers for language
  understanding.
\newblock pages 4171--4186, 2019.

\bibitem{li2021supervision}
Yangguang Li, Feng Liang, Lichen Zhao, Yufeng Cui, Wanli Ouyang, Jing Shao,
  Fengwei Yu, and Junjie Yan.
\newblock Supervision exists everywhere: A data efficient contrastive
  language-image pre-training paradigm.
\newblock In {\em International Conference on Learning Representations}, 2021.

\bibitem{chen2020simple}
Ting Chen, Simon Kornblith, Mohammad Norouzi, and Geoffrey Hinton.
\newblock A simple framework for contrastive learning of visual
  representations.
\newblock In {\em International conference on machine learning}, pages
  1597--1607. PMLR, 2020.

\bibitem{he2020momentum}
Kaiming He, Haoqi Fan, Yuxin Wu, Saining Xie, and Ross Girshick.
\newblock Momentum contrast for unsupervised visual representation learning.
\newblock In {\em Proceedings of the IEEE/CVF conference on computer vision and
  pattern recognition}, pages 9729--9738, 2020.

\bibitem{hjelm2018learning}
R~Devon Hjelm, Alex Fedorov, Samuel Lavoie-Marchildon, Karan Grewal, Phil
  Bachman, Adam Trischler, and Yoshua Bengio.
\newblock Learning deep representations by mutual information estimation and
  maximization.
\newblock In {\em International Conference on Learning Representations}, 2018.

\bibitem{edwards2022translation}
Carl Edwards, Tuan Lai, Kevin Ros, Garrett Honke, and Heng Ji.
\newblock Translation between molecules and natural language.
\newblock {\em arXiv preprint arXiv:2204.11817}, 2022.

\bibitem{edwards2021text2mol}
Carl Edwards, ChengXiang Zhai, and Heng Ji.
\newblock Text2mol: Cross-modal molecule retrieval with natural language
  queries.
\newblock In {\em Proceedings of the 2021 Conference on Empirical Methods in
  Natural Language Processing}, pages 595--607, 2021.

\bibitem{irwin2012zinc}
John~J Irwin, Teague Sterling, Michael~M Mysinger, Erin~S Bolstad, and Ryan~G
  Coleman.
\newblock Zinc: a free tool to discover chemistry for biology.
\newblock {\em Journal of chemical information and modeling}, 52(7):1757--1768,
  2012.

\bibitem{suresh2021adversarial}
Susheel Suresh, Pan Li, Cong Hao, and Jennifer Neville.
\newblock Adversarial graph augmentation to improve graph contrastive learning.
\newblock {\em Advances in Neural Information Processing Systems},
  34:15920--15933, 2021.

\bibitem{gao2022bootstrapping}
Hang Gao, Jiangmeng Li, Wenwen Qiang, Lingyu Si, Changwen Zheng, and Fuchun
  Sun.
\newblock Bootstrapping informative graph augmentation via a meta learning
  approach.
\newblock {\em arXiv preprint arXiv:2201.03812}, 2022.

\bibitem{hu2020pretraining}
Weihua Hu, Bowen Liu, Joseph Gomes, Marinka Zitnik, Percy Liang, Vijay Pande,
  and Jure Leskovec.
\newblock Strategies for pre-training graph neural networks.
\newblock In {\em International Conference on Learning Representations}, 2020.

\bibitem{wu2018moleculenet}
Zhenqin Wu, Bharath Ramsundar, Evan~N Feinberg, Joseph Gomes, Caleb Geniesse,
  Aneesh~S Pappu, Karl Leswing, and Vijay Pande.
\newblock Moleculenet: a benchmark for molecular machine learning.
\newblock {\em Chemical science}, 9(2):513--530, 2018.

\bibitem{martins2012bayesian}
Ines~Filipa Martins, Ana~L Teixeira, Luis Pinheiro, and Andre~O Falcao.
\newblock A bayesian approach to in silico blood-brain barrier penetration
  modeling.
\newblock {\em Journal of chemical information and modeling}, 52(6):1686--1697,
  2012.

\bibitem{tox21}
Tox21 data challenge 2014.
\newblock {\em URL https://tripod.nih.gov/tox21/challenge/}, 2014.

\bibitem{richard2016toxcast}
Ann~M Richard, Richard~S Judson, Keith~A Houck, Christopher~M Grulke, Patra
  Volarath, Inthirany Thillainadarajah, Chihae Yang, James Rathman, Matthew~T
  Martin, John~F Wambaugh, et~al.
\newblock Toxcast chemical landscape: paving the road to 21st century
  toxicology.
\newblock {\em Chemical research in toxicology}, 29(8):1225--1251, 2016.

\bibitem{kuhn2016sider}
Michael Kuhn, Ivica Letunic, Lars~Juhl Jensen, and Peer Bork.
\newblock The sider database of drugs and side effects.
\newblock {\em Nucleic acids research}, 44(D1):D1075--D1079, 2016.

\bibitem{novick2013sweetlead}
Paul~A Novick, Oscar~F Ortiz, Jared Poelman, Amir~Y Abdulhay, and Vijay~S
  Pande.
\newblock Sweetlead: an in silico database of approved drugs, regulated
  chemicals, and herbal isolates for computer-aided drug discovery.
\newblock {\em PloS one}, 8(11):e79568, 2013.

\bibitem{aact}
Aact database.
\newblock {\em URL https://www.ctti-clinicaltrials.org/aact-database}, 2017.

\bibitem{gardiner2011effectiveness}
Eleanor~J Gardiner, Caroline Holliday, John D and{\'\O}D~owd, and Peter
  Willett.
\newblock Effectiveness of 2d fingerprints for scaffold hopping.
\newblock {\em Future medicinal chemistry}, 3(4):405--414, 2011.

\bibitem{hiv}
Aids antiviral screen data.
\newblock {\em URL
  https://wiki.nci.nih.gov/display/NCIDTPdata/AIDS+Antiviral+Screen+Data}.

\bibitem{subramanian2016computational}
Govindan Subramanian, Bharath Ramsundar, Vijay Pande, and Rajiah~Aldrin Denny.
\newblock Computational modeling of $\beta$-secretase 1 (bace-1) inhibitors
  using ligand based approaches.
\newblock {\em Journal of chemical information and modeling},
  56(10):1936--1949, 2016.

\bibitem{velickovic2019deep}
Petar Velickovic, William Fedus, William~L Hamilton, Pietro Li{\`o}, Yoshua
  Bengio, and R~Devon Hjelm.
\newblock Deep graph infomax.
\newblock In {\em International Conference on Learning Representations}, 2019.

\bibitem{hamilton2017inductive}
Will Hamilton, Zhitao Ying, and Jure Leskovec.
\newblock Inductive representation learning on large graphs.
\newblock {\em Advances in neural information processing systems}, 30, 2017.

\bibitem{2008Visualizing}
Van Der~Maaten Laurens and Geoffrey Hinton.
\newblock Visualizing data using t-sne.
\newblock {\em Journal of Machine Learning Research}, 9(2605):2579--2605, 2008.

\bibitem{wu2018unsupervised}
Zhirong Wu, Yuanjun Xiong, Stella~X Yu, and Dahua Lin.
\newblock Unsupervised feature learning via non-parametric instance
  discrimination.
\newblock In {\em Proceedings of the IEEE conference on computer vision and
  pattern recognition}, pages 3733--3742, 2018.

\bibitem{oord2018representation}
Aaron van~den Oord, Yazhe Li, and Oriol Vinyals.
\newblock Representation learning with contrastive predictive coding.
\newblock {\em arXiv preprint arXiv:1807.03748}, 2018.

\bibitem{pytorch}
Adam Paszke, Sam Gross, Francisco Massa, Adam Lerer, James Bradbury, Gregory
  Chanan, Trevor Killeen, Zeming Lin, Natalia Gimelshein, Luca Antiga, et~al.
\newblock Pytorch: An imperative style, high-performance deep learning library.
\newblock In {\em Advances in Neural Information Processing Systems}, pages
  8026--8037, 2019.

\end{thebibliography}
}

\section*{Author contributions}
Ji-Rong Wen managed the study. Bing Su, Dazhao Du, and Ji-Rong Wen designed the methodology of the study. Zhao Yang and Yujie Zhou collected the paired dataset. Dazhao Du performed the pretraining of the model. Dazhao Du, Zhao Yang, Bing Su, and Jiangmeng Li performed experiments on down-stream tasks. Bing Su, Dazhao Du, and Zhao Yang wrote the original draft. All co-authors discussed the results, performed analyses, revised and reviewed the manuscript.

\section*{Corresponding author}
Correspondence to Ji-Rong Wen (jrwen@ruc.edu.cn).

\section*{Competing interests statement}
The authors declare no competing interests.

\end{document}